%% file: main.tex
\documentclass[10pt,twocolumn,letterpaper]{article}

\usepackage[pagenumbers]{cvpr} 


\usepackage{graphicx}
\usepackage{booktabs}

\usepackage[accsupp]{axessibility}  

\usepackage[utf8]{inputenc} 
\usepackage[T1]{fontenc}    
\usepackage{url}            
\usepackage{booktabs}       
\usepackage{amsfonts}       
\usepackage{nicefrac}       
\usepackage{microtype}      
\usepackage{xcolor}         

\usepackage{graphicx}
\usepackage{amsmath}
\usepackage{amssymb}
\usepackage{booktabs}
\usepackage{bbm}
\usepackage{multirow}
\usepackage{textcomp}

\usepackage{makecell}

\definecolor{cvprblue}{rgb}{0.21,0.49,0.74}
\usepackage[pagebackref,breaklinks,colorlinks,allcolors=cvprblue]{hyperref}

\usepackage[capitalize]{cleveref}
\crefname{section}{Sec.}{Secs.}
\Crefname{section}{Section}{Sections}
\Crefname{table}{Table}{Tables}
\crefname{table}{Tab.}{Tabs.}

\def\paperID{9500} 
\def\confName{CVPR}
\def\confYear{2025}

\title{Towards In-the-wild 3D Plane Reconstruction from a Single Image}

\author{Jiachen Liu$^{1}$\thanks{Equal contribution.}
\and
Rui Yu$^{2*}$
\and
Sili Chen$^{3}$
\and
Sharon X. Huang$^{1}$
\and
Hengkai Guo$^{3}$
\and
$^{1}$The Pennsylvania State University
\and
$^{2}$University of Louisville\\
\and
$^{3}$Bytedance\\
}

\begin{document}
\maketitle

\input{sec/abstract}
\input{sec/introduction}
\input{sec/related_work}
\input{sec/datasets}
\input{sec/method}
\input{sec/experiments}
\input{sec/conclusion}
\newpage
{
    \small
    \bibliographystyle{ieeenat_fullname}
    \bibliography{main}
}

\clearpage
\input{sec/X_suppl}


\end{document}

%% file: sec/abstract.tex
\begin{abstract}
3D plane reconstruction from a single image is a crucial yet challenging topic in 3D computer vision. Previous state-of-the-art (SOTA) methods have focused on training their system on a single dataset from either indoor or outdoor domain, limiting their generalizability across diverse testing data. In this work, we introduce a novel framework dubbed \textbf{ZeroPlane}, a Transformer-based model targeting zero-shot 3D plane detection and reconstruction from a single image, over diverse domains and environments. To enable data-driven models across multiple domains, we have curated a large-scale planar benchmark, comprising over 14 datasets and 560,000 high-resolution, dense planar annotations for diverse indoor and outdoor scenes. To address the challenge of achieving desirable planar geometry on multi-dataset training, we propose to disentangle the representation of plane normal and offset, and employ an exemplar-guided, classification-then-regression paradigm to learn plane and offset respectively. Additionally, we employ advanced backbones as image encoder, and present an effective pixel-geometry-enhanced plane embedding module to further facilitate planar reconstruction. Extensive experiments across multiple zero-shot evaluation datasets have demonstrated that our approach significantly outperforms previous methods on both reconstruction accuracy and generalizability, especially over in-the-wild data. Our code and data are available at: \href{https://github.com/jcliu0428/ZeroPlane}{https://github.com/jcliu0428/ZeroPlane}.
\end{abstract}

%% file: sec/introduction.tex
\section{Introduction}
\label{sec:intro}
\begin{figure}[tbp]
    \centering
    \scriptsize
    \setlength{\tabcolsep}{1pt}
    \begin{tabular}{cccccc}
    \centering
        {\includegraphics[width=0.16\linewidth]{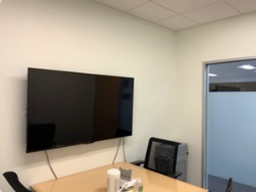}} &  
        {\includegraphics[width=0.16\linewidth]{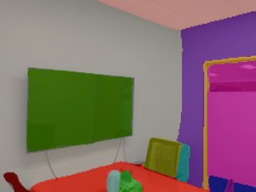}} &  
        {\includegraphics[width=0.16\linewidth]{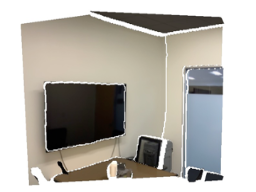}} & 
        {\includegraphics[width=0.16\linewidth]{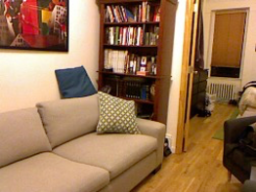}} &
        {\includegraphics[width=0.16\linewidth]{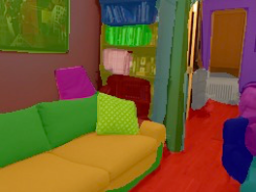}} &
        {\includegraphics[width=0.16\linewidth]{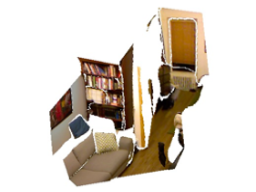}} \\
        \multicolumn{3}{c}{ARKitScenes~\cite{baruch2021arkitscenes}} & \multicolumn{3}{c}{NYUv2~\cite{silberman2012indoor}} \\  

        {\includegraphics[width=0.16\linewidth]{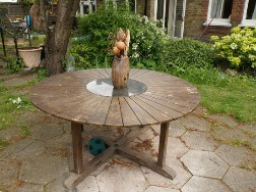}} &  
        {\includegraphics[width=0.16\linewidth]{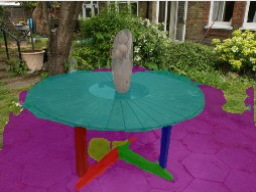}} &  
        {\includegraphics[width=0.16\linewidth]{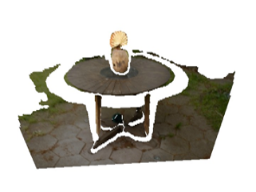}} & 
        {\includegraphics[width=0.16\linewidth]{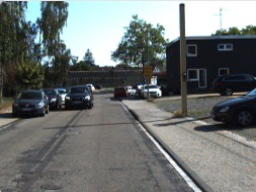}} &
        {\includegraphics[width=0.16\linewidth]{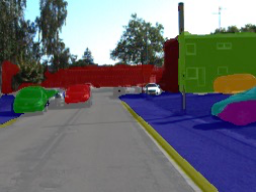}} &
        {\includegraphics[width=0.16\linewidth]{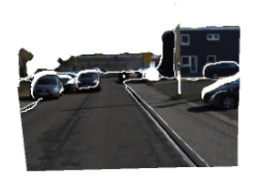}} \\ 
        \multicolumn{3}{c}{Mip-NeRF 360~\cite{barron2022mip}} & \multicolumn{3}{c}{KITTI~\cite{geiger2012we}} \\  

        {\includegraphics[width=0.16\linewidth]{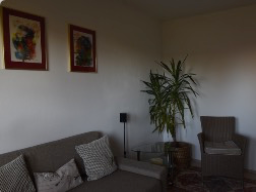}} &  
        {\includegraphics[width=0.16\linewidth]{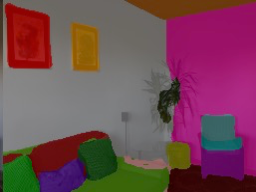}} &  
        {\includegraphics[width=0.16\linewidth]{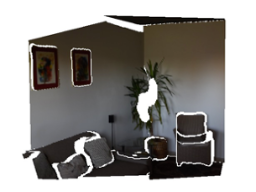}} & 
        {\includegraphics[width=0.16\linewidth]{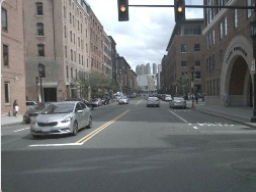}} &
        {\includegraphics[width=0.16\linewidth]{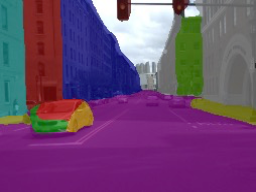}} &
        {\includegraphics[width=0.16\linewidth]{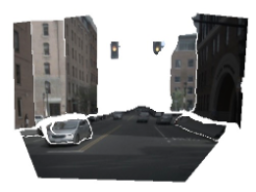}} \\
        \multicolumn{3}{c}{IBims-1~\cite{koch2018evaluation}} & \multicolumn{3}{c}{nuScenes~\cite{caesar2020nuscenes}} \\ 
    \end{tabular}
    \vspace{-2mm}
    \caption{Our plane reconstruction framework, ZeroPlane, demonstrates superior zero-shot generalizability on unseen and even in-the-wild data across diverse indoor and outdoor environments.}
    \vspace{-3mm}
    \label{fig::teaser}
\end{figure}

Recovering 3D geometric primitives from images has been a fundamental yet challenging task in computer vision for decades. Compared with discrete representations such as 3D point clouds, volumetric grids, or meshes, these primitives encapsulate the 3D scene with structural regularities and non-local geometric cues. 3D plane is one of the major geometric primitives with a concise but compact 3D representation and is ubiquitous in diverse man-made environments. 
Perceiving and recovering 3D planes also plays a pivotal role in various applications including augmented reality~(AR)~\cite{chekhlov2007ninja}, localization and mapping~\cite{salas2014dense, hsiao2017keyframe} and robotics~\cite{taguchi2013point}. Extensive studies have been conducted on 3D plane reconstruction from a variety of input data formats such as one or multiple RGB images~\cite{gallup2010piecewise, sinha2009piecewise, liu2018planenet, liu2022planemvs, xie2022planarrecon}, RGB-D sequences~\cite{hsiao2017keyframe} or point cloud~\cite{nan2017polyfit}. Traditional methods have achieved accurate plane recovery by taking specified scene assumptions~\cite{furukawa2009manhattan} or leveraging RGB-D~\cite{salas2014dense, hsiao2017keyframe} or multi-view input~\cite{sinha2009piecewise, gallup2010piecewise}. These approaches involve complex optimization, making them less flexible in heterogeneous environments.
\par
In recent years, data-driven frameworks have emerged to recover 3D planes from a single RGB image with advanced network designs such as CNNs~\cite{liu2018planenet, yang2018recovering, liu2019planercnn, yu2019single} and Transformers~\cite{tan2021planetr, shi2023planerectr}. Despite achieving noteworthy plane reconstruction accuracy on a single indoor~\cite{dai2017scannet} or outdoor dataset~\cite{ros2016synthia}, none of these methods have tested and even considered commendable and generalizable plane reconstruction across diverse environments. Exploring a unified and transferable framework for plane reconstruction from single images across diverse scenes is a promising direction with broad practical applications. For instance, let us consider a scenario where a robot must navigate through both indoor spaces and urban streets. Its ability to perceive its surroundings by identifying and reconstructing 3D planes is essential for effective planning and navigation. An AR user might aim to place virtual objects on diverse identified planar surfaces, such as tabletops in a living room or the ground of a street view. On the other hand, recent breakthroughs~\cite{ranftl2020towards, bhat2023zoedepth, yin2023metric3d, yang2024depth} on zero-shot depth estimation from a single image have validated the feasibility of training a unified model on large-scale data collected from diverse sources, which leads to superior accuracy and generalizability across various testing datasets. This motivates us to explore the development of a 3D plane reconstruction model that generalizes across diverse datasets, particularly for in-the-wild data during inference.
\par
A few indoor datasets~\cite{dai2017scannet, chang2017matterport3d} are benchmarked for previous plane reconstruction methods, whereas the scarcity of high-quality outdoor plane ground-truth labels hinders the possibility to develop a transferrable plane reconstruction system. Moreover, most of existing methods train and test on low-resolution input, limiting the flexibility and quality on plane reconstruction with varied input resolutions. To bridge this gap, in our study, we generate high-resolution plane annotations on several outdoor datasets~\cite{ros2016synthia, gaidon2016virtual, cabon2020virtual, huang2018apolloscape}, and enrich several new indoor benchmarks with plane annotations by leveraging a robust plane estimator and a SOTA pretrained panoptic segmentation model~\cite{cheng2022masked}. Consequently, we create a large-scale plane dataset comprising data from diverse domains and environments. 
\par
In our work, on mixed-dataset plane detection, inspired by the success of prior works~\cite{cheng2022masked, shi2023planerectr, kirillov2023segment}, we employ the advantageous design of Transformer-based detection as our main framework, due to its superior adaptability and scalability for large-scale data training. The key challenge of this problem lies in the plane geometry~(normal and offset parameters) estimation. We empirically find that directly regressing the parameters on mixed training data does not lead to optimal results. Previous SOTA single-dataset plane reconstruction methods~\cite{tan2021planetr, shi2023planerectr} typically represent the plane normal and offset as a coupled, scaled vector, and directly learn to regress it. 
However, planes are distributed in a complex manner, with varied locations, orientations, geometric scales, and appearances across different indoor and outdoor scenes. This diversity brings challenges for the network to accurately regress precise values of a scaled vector. To alleviate this obstacle, we first disentangle the representations of normal and offset, and propose an exemplar-guided, classification-then-regression strategy to learn both components. Moreover, we integrate recent advanced image encoder and pixel-level decoder~\cite{oquab2023dinov2, ranftl2021vision} into our framework, which improves the generalization and robustness of the network on multi-dataset learning. We further design a pixel-geometry-enhanced plane embedding module that encourages the plane queries to exploit useful low-level geometric cues to obtain geometry-enhanced plane embeddings to facilitate the parameter learning. These enhancements help our model perceive distinctive and robust features across various inputs. Our framework has demonstrated consistently accurate performance in plane reconstruction and superior generalizability on various datasets across domains, especially on in-the-wild scenes. Figure~\ref{fig::teaser} shows example results of our model on various unseen datasets.

We summarize our key contributions as follows: 
\begin{itemize}
    \item We introduce the cross-domain plane reconstruction task, and create a large-scale dataset containing high-resolution, dense planar annotations from a variety of indoor and outdoor datasets for benchmarking its training and evaluation.
    \item We present a Transformer-based framework, namely \mbox{ZeroPlane}. To alleviate challenges in learning plane geometry on mixed datasets, we employ a classification-then-regression paradigm for plane normal and offset learning. We further incorporate advanced backbone models to enhance robustness, and propose a pixel-geometry-enhanced module to facilitate plane geometry learning.
    \item Extensive experiments on mixed-dataset training and in-the-wild evaluation demonstrate that our system significantly outperforms existing counterparts in terms of both plane recovery accuracy and generalizability across a variety of indoor and outdoor datasets.
\end{itemize}

%% file: sec/related_work.tex
\section{Related Work}
\label{sec:related_work}
\subsection{3D Plane Reconstruction from Images}

Traditional monocular-based plane reconstruction approaches~\cite{furukawa2009manhattan,gallup2010piecewise,sinha2009piecewise} rely on detecting geometric primitives to reconstruct planar structures with robust estimators such as RANSAC~\cite{fischler1981random}, CRF~\cite{lafferty2001conditional} or MRF~\cite{li1994markov}. Recently, learning-based methods~\cite{liu2018planenet,yang2018recovering,yu2019single,liu2019planercnn,tan2021planetr,qian2020learning} have been proposed, leveraging plane annotations on benchmark datasets~\cite{dai2017scannet, chang2017matterport3d, ros2016synthia}, with a focus of indoor scenes. PlaneNet~\cite{liu2018planenet} pioneered an end-to-end framework to separately learn plane segmentation and geometric parameters, while PlaneRCNN~\cite{liu2019planercnn} used a two-stage network for segmentation and reconstruction. Transformer-based methods, such as PlaneTR~\cite{tan2021planetr} and PlaneRecTR~\cite{shi2023planerectr}, further advanced plane detection and reconstruction.
In outdoor domain, PlaneRecover~\cite{yang2018recovering} detects planes using ground truth depth as supervision. In contrast to these works focusing on single-image plane reconstruction, others~\cite{liu2022planemvs,xie2022planarrecon,agarwala2022planeformers,jin2021planar,tan2023nope,zhang2023structural,ye2023self, watson2024airplanes} address multi-view plane reconstruction.
Although the aforementioned learning-based methods have achieved impressive results on in-distribution data, no existing work offers a single, unified framework for robust 3D plane reconstruction across diverse environments. In this paper, we propose a unified model to achieve consistent 3D plane reconstruction across varied environments.

\subsection{Zero-shot Single-Image 3D Reconstruction}

Zero-shot single-image 3D reconstruction aims to generate 3D structures from a single image of out-of-domain data without fine-tuning. To achieve generalization to unseen data, previous research~\cite{ranftl2020towards,antequera2020mapillary,ranftl2021vision,eftekhar2021omnidata,yin2022towards} has trained models on mixed large-scale datasets and introduced diverse geometry encoding approaches. For monocular depth estimation, MiDaS~\cite{ranftl2020towards} combines five datasets and over two million images, employing scale-invariant loss for robust training, a method further explored in Omnidata~\cite{eftekhar2021omnidata}. However, these frameworks are limited to predicting depth only up to an unknown scale. Recent approaches~\cite{guizilini2023towards,yin2023metric3d,bhat2023zoedepth} expand this to metric scale; for instance, ZeroDepth~\cite{guizilini2023towards} leverages geometric embeddings for metric depth estimation.
Depth Anything~\cite{yang2024depth,yang2024depth2} introduced a foundation model trained on 62M labeled and unlabeled images for monocular depth estimation. 
For single-view object reconstruction, zero-shot approaches~\cite{liu2023zero,weng2023zeroavatar,sargent2023zeronvs,huang2023zeroshape} also exhibit strong generalization on out-of-domain or in-the-wild images.
Despite these advancements, domain-generalized 3D plane reconstruction remains unexplored. To the best of our knowledge, our work represents the pioneering exploration of zero-shot 3D plane reconstruction from a single image.

%% file: sec/datasets.tex
\setlength{\tabcolsep}{3pt}
\begin{table}[t!]
\caption{Statistics of the datasets used in our work. Top: Datasets used for training and validation. Bottom: Datasets used for zero-shot evaluation.}
\label{tab::dataset}
\vspace{-3mm}
\centering
\resizebox{1.0\linewidth}{!}{
\begin{tabular}{ccccccc}
\toprule
\centering
Dataset & Indoor & Outdoor & Label & \makecell{\#Training \\ Images} & \makecell{\#Validation \\ Images} &~\#Planes \\
\midrule
ScanNetv1~\cite{dai2017scannet} & \checkmark & & RGB-D & 47923 & 719 & 281268\\
ScanNetv2~\cite{dai2017scannet} & \checkmark & & RGB-D & 140773 & 859 & 1445183\\
Matterport3D~\cite{chang2017matterport3d} & \checkmark & & RGB-D & 20854 & 943 & 227157\\
Replica~\cite{straub2019replica} & \checkmark & & Synthetic & 5533 & 400 & 90466\\
HM3D~\cite{yadav2023habitat} & \checkmark & & Synthetic & 79994 & 2000 & 1298635\\
DIODE~(Indoor Split)~\cite{vasiljevic2019diode} & \checkmark & & RGB-D & 7350 & 731 & 37447\\
Taskonomy~\cite{zamir2018taskonomy} & \checkmark & & RGB-D & 123740 & 499 & 1450599\\
Synthia~\cite{ros2016synthia} & & \checkmark & Synthetic & 24636 & 999 & 143698\\
Virtual KITTI~\cite{gaidon2016virtual,cabon2020virtual} & & \checkmark & Synthetic & 17720 & 970 & 71171\\
Sanpo~(Synthetic Split)~\cite{waghmare2023sanpo} & & \checkmark & Synthetic & 89854 & 488 & 647530\\
\midrule
NYUv2~\cite{silberman2012indoor} & \checkmark & & RGB-D & - & 654 & 4849\\
7-Scenes~\cite{shotton2013scene} & \checkmark & & RGB-D & - & 758 & 5433\\
ApolloScape~\cite{huang2018apolloscape}~(Stereo Split) & & \checkmark & Stereo & - & 999 & 4241\\
ParallelDomain~\cite{guizilini2021geometric, qian2020learning} & & \checkmark & Synthetic & - & 356 & 2350\\
\bottomrule
\end{tabular}}
\vspace{-5mm}
\end{table}

\section{Datasets}
\label{sec:datasets}
In this section, we present the benchmark datasets we have used in this paper for training and zero-shot evaluation, and introduce the way we generate high-quality groundtruth plane annotations for the newly-adopted datasets.
\subsection{Dataset Collection}
To achieve our goal of building a transferrable 3D plane reconstruction system, we need a scale-considerable, densely-annotated planar benchmark dataset sourced from diverse environments. 
While prior studies have implemented plane fitting and annotation pipelines for indoor environments using a few large-scale semantic RGB-D benchmarks~\cite{dai2017scannet, chang2017matterport3d}, the limited data diversity has hindered the model's ability to generalize to new, unseen testing data.
On outdoor scenes, to the best of our knowledge, there is no existing dataset that contains dense planar annotations in outdoor environments and enables the training of a planar reconstruction model.
Although PlaneRecover~\cite{yang2018recovering} has used Synthia~\cite{ros2016synthia}, there are no large-scale plane mask labels used during training and authors manually annotate $100$ images with dense plane masks for evaluation purpose. 
To tackle the challenge of limited diverse and high-quality planar ground truth data, we expanded our approach by adopting more indoor datasets and generating planar ground truth for a few outdoor datasets which are sampled from a variety of environments. 
\par
Specifically, for indoor data, we first adopt existing datasets with plane annotations, such as ScanNet~\cite{dai2017scannet}~(referred to as ScanNetv1) with plane labels generated in~\cite{liu2018planenet}, a larger-scale version~(ScanNetv2) produced by~\cite{liu2019planercnn}, Matterport3D~\cite{chang2017matterport3d} annotated by~\cite{jin2021planar}, and NYUv2~\cite{silberman2012indoor} annotated by~\cite{yu2019single}. Additionally, we generate plane labels for several new indoor benchmarks, including two photo-realistic datasets: Replica~\cite{straub2019replica} and Habitat-Matterport3D~\cite{yadav2023habitat}, as well as three real RGB-D datasets: 7-Scenes~\cite{shotton2013scene}, DIODE~\cite{vasiljevic2019diode}, and Taskonomy~\cite{zamir2018taskonomy}. 
For outdoor datasets, the impracticability of obtaining accurate dense depth maps in the real world poses challenges to fitting planes and producing dense plane annotations. 
To this end, we leverage synthetic datasets containing accurate and complete depth maps including Synthia~\cite{ros2016synthia}, Virtual KITTI~\cite{gaidon2016virtual, cabon2020virtual}, Sanpo~\cite{waghmare2023sanpo}, ParallelDomain~\cite{guizilini2021geometric, qian2020learning}, as well as ApolloScape~\cite{huang2018apolloscape} with dense disparity map from stereo camera, to generate planar ground truth. We will introduce the details of the dense plane annotation procedure in Sec.~\ref{dataset_anno}.

In our setup, we utilize NYUv2 and 7-Scenes for zero-shot indoor evaluation, while ApolloScape and ParallelDomain serve as the outdoor zero-shot evaluation datasets. 
The remaining datasets are employed for mixed-domain training and validation. Table~\ref{tab::dataset} presents the statistics of each dataset. Please refer to the supplementary material for further details on the datasets utilized.
\subsection{Plane Label Generation for New Datasets}
\label{dataset_anno}
Previous works~\cite{liu2018planenet, liu2019planercnn} perform plane annotation by fitting planes on the mesh of the entire scene using RANSAC~\cite{fischler1981random}, and then render the obtained 3D planes back to every image. Although this approach may produce more complete plane labels, we empirically find that it is challenging to obtain clean and accurate mesh from image sequences through TSDF-Fusion~\cite{zeng20173dmatch} in outdoor scenes. On the other hand, such annotation pipeline cannot work if a dataset does not provide image sequences. To address the issue and improve the generation flexibility, we fit planes for every single image onto the point cloud back-projected from the depth map on both indoor and outdoor new datasets. This approach makes the label annotation more efficient and flexible compared to mesh-based fitting, without sacrificing label precision.
\par
Directly fitting planes onto the entire point cloud remains challenging and computationally demanding. Previous methods fit planes onto the points of each object instance where the semantic labels are from the datasets(\eg, ScanNet, Matterport3D). 
For newly adopted datasets like Replica, HM3D, and Sanpo, which offer object-level semantic ground truths, we directly incorporate their annotations into our annotation pipeline. For datasets lacking complete semantic labels such as DIODE, Taskonomy, 7-Scenes, Synthia, Virtual KITTI, ApolloScape, and ParallelDomain, we employ a SOTA image segmentation network Mask2Former~\cite{cheng2022masked} to obtain their panoptic segmentation results as pseudo ground truth. Subsequently, on the images for each dataset, we sequentially fit planes and annotate planar masks onto the back projected points of each background stuff class or foreground instance. 
\par
Although most of prior works train and test their methods on low-resolution~($256\times192$) images, we claim that high-resolution dataset is a pre-requisite for high-quality plane reconstruction. To this end, we generate groundtruth planes with $640\times480$ resolution for all datasets. More details on the visualization and quality measurement of plane groundtruth generation can be found in our supplementary materials.


%% file: sec/method.tex
\definecolor{darkgreen}{rgb}{0,0.6,0}

\begin{figure*}[t]
	\centering
	\includegraphics[width=0.85\linewidth]{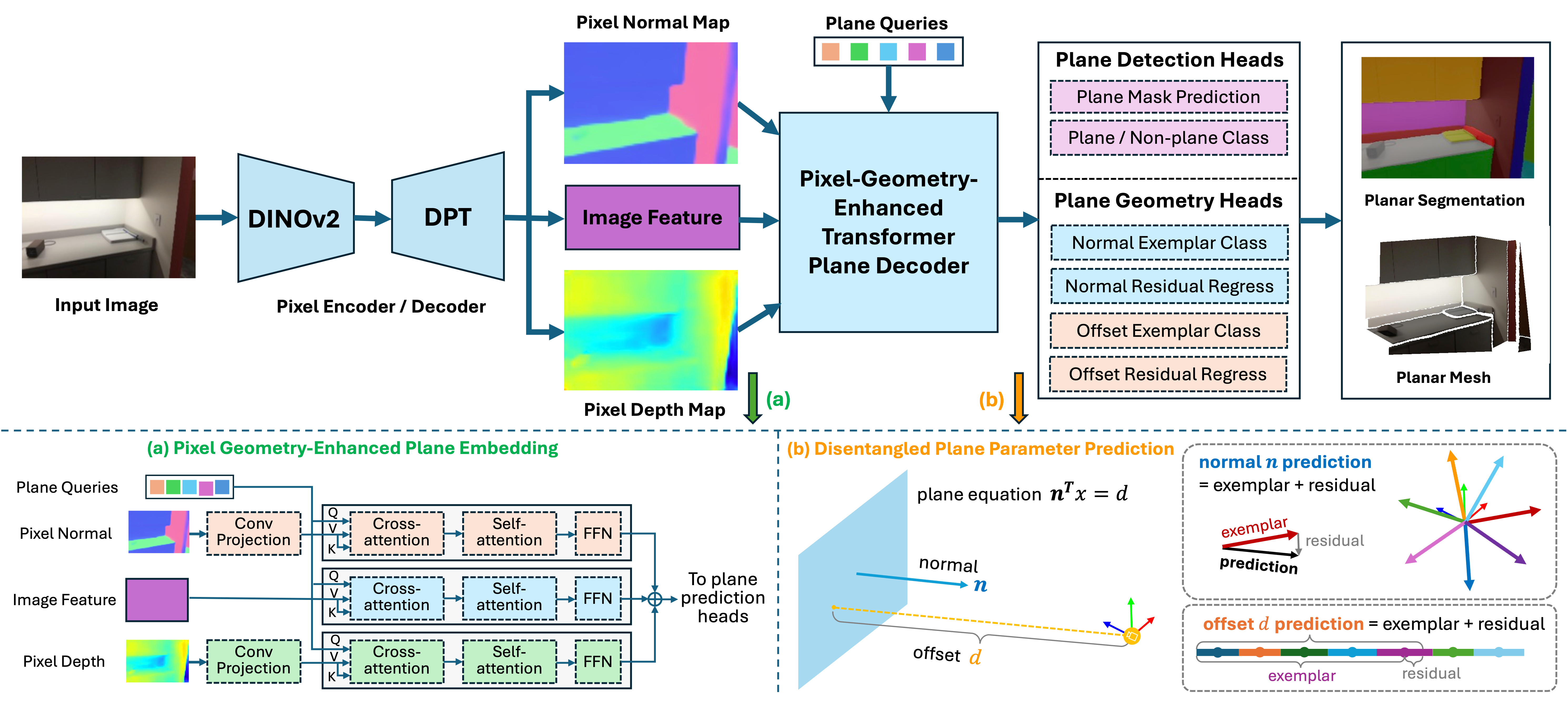}
	\caption{Our proposed \textbf{ZeroPlane} framework. Taking a single image as input, our model first extracts image features from encoder and decoder networks. The plane queries and the predicted pixel-level depth and normal map serve as inputs to module \textcolor{darkgreen}{(a)} to obtain geometry-enhanced plane embeddings. These embeddings are then fed into plane detection heads for mask and classification predictions, and plane geometry heads for disentangled plane normal and offset predictions. Notably, the normal and offset are both learned via a classification-then-regression paradigm, as illustrated in \textcolor{orange}{(b)}.
    Each predicted plane corresponds to a plane query, with all queries sharing the same plane prediction heads.}
    \label{fig:framework}
\end{figure*}

\section{Methodology}
\label{sec:method}
In this section, we present the overall design of our transformer-based model (Sec.~\ref{sec:overall_framework}), followed by our disentangled, classification-then-regression schemes for plane normal and offset (Sec.~\ref{sec: disentangled_nd}), and the pixel geometry enhanced plane embedding module (Sec.~\ref{sec: geo_attn}). Lastly, we describe the loss functions we have used (Sec.~\ref{sec: loss}).

\subsection{Transformer-based Plane Reconstruction Framework}
\label{sec:overall_framework}
To develop a unified, generalizable, and scalable plane reconstruction system, our model is designed and built upon the SOTA Transformer-based detection and segmentation frameworks~\cite{carion2020end, cheng2022masked}, which have been employed in various 3D geometry prediction tasks~\cite{yang2024polymax, shi2023planerectr}.
Planes are treated as query embeddings and detected through a query-based reasoning approach in this framework.
As shown in Fig.~\ref{fig:framework}, the main architecture of our framework follows the overall design of Mask2Former~\cite{cheng2022masked} and PlaneRecTR~\cite{shi2023planerectr}, consisting of a backbone encoder network which extracts multi-scale feature maps and a Transformer-based decoder. Within the decoder, a set of learnable queries are interleaved with multi-scale features from the encoder, performing multi-layer alternated cross-attention and self-attention to produce optimized instance-wise plane embeddings. The embeddings are then processed by separate projection heads to predict plane-wise outputs respectively.

\noindent\textbf{Pixel-level encoder and decoder.} 
Our goal is to consistently generate robust representations on diverse data through a distinctive encoder. To this end, unlike PlaneRecTR which uses a pretrained SwinTransformer~\cite{liu2021swin}, in our default setting, we integrate the pretrained DINOv2~\cite{oquab2023dinov2} model as our encoder, which is a more powerful ViT~\cite{dosovitskiy2020image} trained on millions of unlabeled samples via contrastive learning, to extract multi-scale feature embeddings.
To obtain pixel-level representations which are efficient on geometric prediction tasks, we employ the RefineNet-based~\cite{lin2017refinenet} fusion blocks used in DPT~\cite{ranftl2021vision} as our pixel decoder. These blocks integrate the embeddings from DINOv2 at different resolutions as input, leading to decoded multi-scale image pixel-wise feature maps $\mathbf{F}$.

\noindent\textbf{Transformer decoder for plane instance-wise prediction.} 
Following PlaneRecTR~\cite{shi2023planerectr}, our model utilizes the learnable query-based reasoning scheme to learn various plane-level outputs. Within the Transformer decoder, the embeddings of plane queries perform masked cross-attention with the encoded image feature maps $\mathbf{F}$, followed by self-attention among queries and feed-forward networks (FFN) for rounds of optimization. The optimized query embeddings are then projected by separate heads to derive instance-level planar predictions. These predictions encompass plane geometry attributes (i) plane normal $\mathbf{n}$ (3D orientation) and (ii) plane offset $d$ (distance from camera to plane), alongside plane detection attributes including (iii) classification score $C$ (probability of belonging to a plane or not), as well as (iv) plane 2D segmentation mask logits $\mathbf{M}\in\mathbb{R}^{H\times W}$. With the predicted plane masks and parameters, the planar depth map can be recovered via $D_{M} = d~/~(n^{T}\cdot K^{-1}q)$, where $K$ denotes camera intrinsics and $q$ denotes pixel homogeneous coordinate and $D_{M}$ denotes depth over planar mask pixels.

\subsection{Plane Normal and Offset Estimation for Multi-dataset Training}
\label{sec: disentangled_nd}
\noindent\textbf{Decoupled representation of plane normal and offset.} 
Current SOTA Transformer-based frameworks for plane reconstruction~\cite{tan2021planetr,shi2023planerectr} represent the plane parameter as the division of normal and offset ($\mathbf{n}/d$) and learn to regress it directly. This representation greatly couples the normal and offset. However, they hold distinct geometric properties: the normal indicates the 3D plane orientation, while the offset denotes the distance from the plane to the camera. In our experiments, we discover that this coupling does not affect single-dataset training and evaluation significantly, as the geometric scale of the training data remains relatively consistent. In contrast, in cross-dataset training with higher geometric range variation (\eg, mixed interior scenes and street views during training), this coupled representation can lead to challenges in regressing the precise value consistently regarding varied inputs for the network. Thus, we argue that the two parameters should be disentangled. Inspired by this, we set two heads after the decoder to learn plane-level normal and offset separately.

\noindent\textbf{Classification-then-regression learning scheme.}
We empirically observed that directly regressing disentangled values of normal and offset does not yield the desired accuracy. One significant factor contributing to this could be the notable disparity between the distribution of plane parameters in indoor and outdoor domains. The varied appearances and geometric scales of the planes can present significant obstacles when the network attempts to directly predict normal and offset values.
To address the challenges arising from domain gaps, inspired by prior works~\cite{liu2019planercnn, bhat2021adabins, fu2018deep, liu2022planemvs, xie2022planarrecon} that employs bin-based or cluster-based classification for geometric prediction tasks, we regard both plane normal and offset predictions as classification-then-regression tasks. Our rationale is that, it is less demanding for the network to learn to classify the orientation and the distance scale to one of the ``exemplars''(cluster centers) than directly regress the exact values. Although PlaneRCNN employs a similar paradigm on normal estimation, they focus on single-dataset training, and predict the offset through normal and depth map. This is different from our classification-then-regression strategy on offset. We leverage ground-truth plane parameters from mixed large-scale training data to derive a set of normal and offset exemplars through clustering algorithms~(K-Means in our case). As illustrated in Fig.~\ref{fig:framework}(b), this clustering process yields $K_{n}$ normal exemplars $\mathbf{\hat{n}}\in \mathbb{R}^{K_{n}\times 3}$ and $K_{d}$ offset exemplars $\hat{d}\in \mathbb{R}^{K_{d}\times 1}$, representing widely-distributed planes across scenes and encapsulating geometric priors. We employ two MLP decoders as the classification heads to learn the normal class $C_{n}$ and offset class $C_{d}$ for each plane instance. We utilize another two MLPs to predict the normal residual $\mathbf{r_n}\in \mathbb{R}^{K_{n}\times 3}$ and offset residual $r_{d}\in \mathbb{R}^{K_{d}\times 1}$ \wrt each exemplar. Finally, the plane normal and offset sum the predicted (assuming $i$-th normal and $j$-th offset) exemplar and the residual vector or value as the prediction.
\begin{equation}
    \mathbf{n} = \mathbf{\hat{n}}^{(i)} + \mathbf{r_n}^{(i)}, \quad d = \hat{d}^{(j)} + r_d^{(j)}
\end{equation}

\subsection{Pixel-Geometry-Enhanced Plane Embedding}
\label{sec: geo_attn}

\noindent\textbf{Pixel-level depth and normal prediction as auxiliary tasks.}
Transformer-based plane detectors~\cite{tan2021planetr,shi2023planerectr} have excelled in capturing high-level planar \textit{semantic} information. However, incorporating low-level \textit{geometric} cues such as pixel depth and normal is profitable for plane identification and geometry recovery. Pixel-level depth indicates the distance from the surface point to the camera, while pixel-level surface normal provides useful cues for identifying plane orientation and boundaries. Motivated by this, we propose integrating pixel-level depth and normal estimation as auxiliary tasks to enable the network to learn the overall geometry of the scene and leverage valuable geometric cues. As depicted in Fig.~\ref{fig:framework}(a), following the encoder, we incorporate two CNN blocks to predict pixel depth map $\mathbf{D}\in \mathbb{R}^{H\times W}$ and normal map $\mathbf{N}\in \mathbb{R}^{H\times W\times 3}$, respectively.

\noindent\textbf{Geometry-enhanced plane embedding.}
We found that incorporating the pixel depth and normal training merely as auxiliary tasks did not yield substantial improvement. While the encoder and decoder might implicitly learn geometric information through multi-task training, the plane queries are unable to leverage the low-level geometric cues to enhance instance-wise predictions. To address this bottleneck, we first project the depth and normal into the embedding space, denoted as $\mathbf{F_{D}}$ and $\mathbf{F_{N}}$ through separate CNN layers.
Then the plane query embeddings $\mathbf{Q}$ perform alternating cross- and self-attentions with the projected geometric features $\mathbf{F_{D}}$ and $\mathbf{F_{N}}$ in a similar way as that with image features, resulting in enhanced embeddings $\mathbf{X_{D}}$ and $\mathbf{X_{N}}$:
\begin{equation}
    \mathbf{X_{D}} = Attn(\mathbf{Q}, \mathbf{F_{D}}), \quad \mathbf{X_{N}} = Attn(\mathbf{Q}, \mathbf{F_{N}})
\end{equation}
The final geometric-enhanced embedding $\mathbf{X}$ is represented as the sum of the original embedding $\mathbf{X_{F}} = Attn(\mathbf{Q}, \mathbf{F})$ and the above two geometric embeddings: $\mathbf{X} = \mathbf{X_{F}} + \mathbf{X_{D}} + \mathbf{X_{N}}$.
The attention mechanism guides the plane queries to discover fine-grained context geometric cues (\eg, plane boundaries) from the pixel-level geometric predictions. Leveraging geometric attention modules, the decoder can extract geometry-enhanced plane embeddings, which proved to be beneficial for decoding various outputs during plane reconstruction.

\subsection{Loss Functions}
\label{sec: loss}
We employ the bipartite matching strategy, as used in previous works~\cite{carion2020end, cheng2022masked, shi2023planerectr}, to match predictions and ground-truth plane instances during training. Our overall loss function includes the following components: plane classification loss $L_{c}$, mask loss $L_{m}$~(joint cross entropy and dice loss), normal classification loss $L_{n_{c}}$ and residual regression loss $L_{n_{r}}$ ($L_{1}$ loss), offset classification loss $L_{d_{c}}$ and residual regression loss $L_{d_{r}}$ ($L_{1}$ loss), pixel-level depth loss $L_{p_{d}}$ ($L_{1}$ loss), and normal loss $L_{p_{n}}$ (joint $L_{1}$ and cosine distance). $L_{p_{d}}$ is only supervised on the pixels with valid ground-truth depth values, while $L_{p_{n}}$ is only supervised on the ground-truth planar pixels. For residual supervision, we leverage the ground-truth class to index the predicted residuals, ensuring only the residuals of selected classes are supervised during training. Cross-entropy loss is employed for each classification task. Note that following the design of~\cite{cheng2022masked}, we supervise the predictions from every layer of the transformer decoder during training. The final loss is computed as:
\begin{equation}
\begin{aligned}
    L = \lambda_{c}L_{c} +~\lambda_{m}L_{m} +~\lambda_{n_{c}}L_{n_{c}} +~\lambda_{n_{r}}L_{n_{r}} \\
    +~\lambda_{d_{c}}L_{d_{c}} +~\lambda_{d_{r}}L_{d_{r}} +~\lambda_{p_{d}}L_{p_{d}} +~\lambda_{p_{n}}L_{p_{n}},
\end{aligned}
\end{equation}
where $\lambda_{*}$ are the corresponding loss weights.


%% file: sec/experiments.tex
\section{Experiments}
\label{sec:experiments}
\subsection{Implementation Details}
We have implemented our framework using PyTorch based on the code of~\cite{cheng2022masked, shi2023planerectr}. 
Our model initializes the encoder weights from a pretrained DINOv2-base~\cite{oquab2023dinov2} model and is trained in an end-to-end fashion with a batch size of 16 using the AdamW~\cite{loshchilov2017decoupled} optimizer for $50$K steps.
The learning rate is set to $1\times 10^{-4}$, with a $10\times$ decay at $40K$ and $47K$ steps.

For normal exemplar clustering, we set $K_{n} = 7$ and apply the K-Means algorithm to the entire mix-training dataset. 
For offset exemplar clustering, to avoid the clusters being dominated by a single domain due to the significant geometric scale disparity between indoor and outdoor data, we initially divide all planes into two groups based on their offset values using a threshold of $20m$. 
We then separately cluster $10$ exemplars for each group and merge the resulting exemplars, resulting in a total of $K_{d}=20$ exemplars.
During inference, we essentially follows the manner of PlaneRecTR~\cite{shi2023planerectr}, preserving the plane queries classified as planes and decoding its plane masks and parameters. Then planar depth map can be reconstructed.
For further experimental and network architecture details, please refer to the supplementary material.

\subsection{Evaluation Metrics}
Following previous works~\cite{liu2018planenet, tan2021planetr, shi2023planerectr} on single-view plane reconstruction, we evaluate plane detection quality using several segmentation metrics including variation of information (VOI), rand index (RI), and segmentation covering (SC).
Geometric reconstruction accuracy is evaluated by computing the average plane recall under various depth and normal error thresholds.
Specifically, we employ depth thresholds of 0.05m/0.1m/0.6m for indoor datasets and 1m/3m/10m for outdoor datasets.
Normal thresholds of 5°/10°/30° are measured on both domains.

\setlength{\tabcolsep}{6pt}
\begin{table*}[t!]
\caption{Zero-shot evaluation of different methods or settings on indoor datasets (NYUv2~\cite{silberman2012indoor}, 7-Scenes~\cite{shotton2013scene}) and outdoor datasets (ParallelDomain~\cite{guizilini2021geometric, qian2020learning}, ApolloScape~\cite{huang2018apolloscape}). (S: trained on ScanNetv1~\cite{dai2017scannet}; S-v2: trained on ScanNetv2, whose training set is much larger than ScanNetv1; M: trained on mixed datasets.)}
\label{tab::mixed}
\centering
\resizebox{0.9\linewidth}{!}{
\begin{tabular}{c|c|ccc|ccc|ccc}
\hline
Evaluation Dataset & \multirow{2}{*}{Method} & \multicolumn{3}{|c}{Plane Segmentation} & \multicolumn{3}{|c}{Plane Recall (depth)} & \multicolumn{3}{|c}{Plane Recall (normal)} \\ 
\cline{3-11}
(Indoor) &   & RI(↑) & VOI(↓)    & SC(↑)  & @0.05m     & @0.1m   & @0.6m & @5° & @10° & @30°   \\ 
\hline
\multirow{9}{*}{NYUv2~\cite{silberman2012indoor}} & \multicolumn{1}{l|}{PlaneRCNN~({S-v2})~\cite{liu2019planercnn}} & 0.84 & 1.60 & 0.61 & 3.04 & 9.82 & \textbf{43.41} & 9.24 & 24.12 & 43.60 \\
& \multicolumn{1}{l|}{PlaneAE~({S})~\cite{yu2019single}} & 0.89 & 1.40 & 0.68 & 1.26 & 3.36 & 15.06 & 11.18 & 26.54 & 38.52 \\
& \multicolumn{1}{l|}{PlaneTR~({S})~\cite{tan2021planetr}} & 0.89 & 1.16 & 0.71 & 2.14 & 6.33 & 28.30 & 12.42 & 27.01 & 34.63 \\
& \multicolumn{1}{l|}{PlaneRecTR~(S)~\cite{shi2023planerectr}} & 
\textbf{0.92} & 1.05 & \textbf{0.75} & 3.3 & 10.13 & 41.47 & 16.42 & 34.42 & 45.7 \\
& \multicolumn{1}{l|}{Ours-DINO-B~(S)} & \textbf{0.92} & \textbf{1.02} & \textbf{0.75} & \textbf{4.33} & \textbf{11.03} & 41.45 & \textbf{24.05} & \textbf{39.88} & \textbf{47.35} \\
\cline{2-11}
& \multicolumn{1}{l|}{PlaneRecTR~(M)~\cite{shi2023planerectr}} & 0.91 & 1.01  & 0.73 & 5.77 & 14.29 & 52.01 & 24.97 & 40.96 & 54.71\\
& \multicolumn{1}{l|}{Ours-DINO-B~(M)} & \textbf{0.92} & \textbf{0.93} & \textbf{0.75} & 8.54 & 17.86 & 55.08 & 37.29 & 47.58 & 57.19 \\
& \multicolumn{1}{l|}{Ours-DINO-L~(M)} & \textbf{0.92} & 0.96 & \textbf{0.75} & 9.32 & 20.29 & 55.37 & 38.15 & \textbf{47.76} & \textbf{56.3} \\
& \multicolumn{1}{l|}{Ours-Dust3R~(M)} & \textbf{0.92} & 0.97 & \textbf{0.75} & \textbf{9.94} & \textbf{21.2} & \textbf{56.03} & \textbf{38.32} & 47.6 & 56.14 \\
\hline
\multirow{4}{*}{7-Scenes~\cite{shotton2013scene}} & \multicolumn{1}{l|}{PlaneRecTR~(M)~\cite{shi2023planerectr}} & \textbf{0.92} & \textbf{1.04} & 0.78 & 3.20 & 10.97 & 45.30 & 11.17 & 26.67 & 42.17 \\
& \multicolumn{1}{l|}{Ours-DINO-B~(M)} & \textbf{0.92} & 1.06 & \textbf{0.79} & 8.19 & 17.19 & 48.13 & 23.96 & 35.43 & 44.17\\
& \multicolumn{1}{l|}{Ours-DINO-L~(M)} & \textbf{0.92} & \textbf{1.04}  & 0.78 & 7.16 & 16.25 & 48.74 & 25.51 & \textbf{36.32} & \textbf{44.49} \\
& \multicolumn{1}{l|}{Ours-Dust3R~(M)} & \textbf{0.92} & 1.08  & 0.78 & \textbf{9.11} & \textbf{19.14} & \textbf{49.73} & \textbf{26.69} & 36.11 & 44.43 \\
\hline
\hline
Evaluation Dataset & \multirow{2}{*}{Method} & \multicolumn{3}{c}{Plane Segmentation} & \multicolumn{3}{|c}{Plane Recall (depth)} & \multicolumn{3}{|c}{Plane Recall (normal)} \\ 
\cline{3-11}
(Outdoor) &  & RI(↑) & VOI(↓)  & SC(↑)  & @1m     & @3m   & @10m & @5° & @10° & @30°   \\ 
\hline
\multirow{4}{*}{ParallelDomain~\cite{guizilini2021geometric, qian2020learning}} & \multicolumn{1}{l|}{PlaneRecTR~(M)~\cite{shi2023planerectr}} & 0.92 & 0.96 & 0.76 & 19.11 & 35.49 & 51.92 & 24.17 & 35.11 & 53.36\\
& \multicolumn{1}{l|}{Ours-DINO-B~(M)} & 0.94 & 0.65 & 0.85 & 25.96 & 44.51 & 63.45 & 53.23 & 65.74 & 69.62 \\
& \multicolumn{1}{l|}{Ours-DINO-L~(M)} & \textbf{0.95} & \textbf{0.57} & \textbf{0.87} & \textbf{27.66} & 44.72 & \textbf{64.0} & 52.47 & \textbf{66.13} & \textbf{70.47} \\
& \multicolumn{1}{l|}{Ours-Dust3R~(M)} & 0.94 & 0.69 & 0.84 & 27.4 & \textbf{46.26} & 63.19 & \textbf{53.91} & 65.36 & 69.87 \\
\hline
\multirow{4}{*}{ApolloScape~\cite{huang2018apolloscape}} & \multicolumn{1}{l|}{PlaneRecTR~(M)~\cite{shi2023planerectr}} & 0.94 & 0.38 & 0.91 & 10.09 & 43.06 & 57.06 & 22.99 & 33.32 & 51.88 \\
& \multicolumn{1}{l|}{Ours-DINO-B~(M)} & \textbf{0.96} & \textbf{0.3} & \textbf{0.93} & 10.28 & 40.44 & 57.34 & 27.56 & 39.78 & 56.05 \\
& \multicolumn{1}{l|}{Ours-DINO-L~(M)} & \textbf{0.96} & \textbf{0.3}  & \textbf{0.93} & \textbf{12.26} & \textbf{46.4} & \textbf{62.08} & \textbf{28.51} & \textbf{40.32} & \textbf{59.87} \\
& \multicolumn{1}{l|}{Ours-Dust3R~(M)} & 0.95 & 0.35  & 0.92 & 10.96 & 44.21 & 58.9 & 24.43 & 37.3 & 55.15 \\
\hline
\end{tabular}
}
\end{table*}


\subsection{Quantitative and Qualitative Evaluation} 
\noindent\textbf{Zero-shot evaluation.} To demonstrate the superior generalizability of our framework, we first conduct zero-shot evaluation on datasets from different domains that were unseen during training, as shown in Table~\ref{tab::mixed}. Specifically, on NYUv2, we evaluate the methods solely trained on ScanNet for fair comparison. Our approach significantly outperforms other SOTA methods across almost all metrics in terms of plane segmentation and 3D geometry. 
Moreover, when trained on multiple indoor and outdoor datasets, our method exhibits substantial improvement on different benchmark datasets, both indoor (NYUv2~\cite{silberman2012indoor}, 7-Scenes~\cite{shotton2013scene}) and outdoor (ParallelDomain~\cite{guizilini2021geometric, qian2020learning}, ApolloScape~\cite{huang2018apolloscape}), compared to mixed-trained counterpart PlaneRecTR (which utilizes the same training setting as ours).
This validates the superiority of our network architecture design and the disentangled classification-then-regression learning paradigm. To validate our versability on feature encoders, we further adopt DINOv2 trained with ViT-Large architecture~(DINO-L) and a recent geometric 3D foundation model Dust3R~\cite{wang2024dust3r} as our backbone network as a stronger variant. One can see that, our model has generally shown robustness over the choices and have further achieved performance gain from the prior knowledge from these models which has pretrained on millions of versatile images.

\noindent\textbf{Ablation study.} Under mixed dataset training setting, we conduct an ablation study on each proposed component, including the use of DINO-Base encoder and DPT decoder~(short as DPT), the classification-then-regression learning scheme~(Cls-Reg), and the attention-based geometry-enhanced plane embedding module~(Geo-Attn). As shown in Table~\ref{tab::ablation}, the first row denotes the base version of our method that employs the original backbone~(SwinTransformer encoder and decoder) and directly regresses $n/d$. To demonstrate the unique benefits of our proposed modules on \textit{multi-dataset training}, we further conduct the ablation study on single-dataset~(ScanNet) training in the same manner. One can observe that under mixed-dataset setting, each component plays an remarkable role to the plane reconstruction result, significantly outperforming the improvement of which brings on single-dataset training. This validates the notable effectiveness of our employed modules or training strategies on multi-domain scenario.

\begin{table}[t!]
\caption{Ablation studies on the contributed components under both single-dataset (ScanNet) training and mixed-dataset training schemes, evaluated on the NYUv2 dataset.}
\label{tab::ablation}
\vspace{-2mm}
\centering
\resizebox{1.0\linewidth}{!}{
\begin{tabular}{c|ccc|ccc|ccc}
\hline
\multirow{2}{*}{Training datasets} & \multicolumn{3}{c}{Component} & \multicolumn{3}{|c}{Plane Recall (depth)} & \multicolumn{3}{|c}{Plane Recall (normal)} \\ 
\cline{5-10}
& DPT & Cls-Reg & Geo-Attn & @0.05m & @0.1m & @0.6m & @5° & @10° & @30° \\ 
\hline
\multirow{4}{*}{ScanNet} & - & - & - & 2.76 & 8.21 & 39.37 & 10.87 & 29.61 & 45.1 \\
& \checkmark & - & - & 4.17 & 10.66 & 41.2 & 19.78 & 37.93 & 47.0 \\
& \checkmark & \checkmark & - & 3.88 & 9.71 & 41.27 & \textbf{24.27} & 38.73 & 45.84 \\
 
& \checkmark & \checkmark & \checkmark & \textbf{4.33} & \textbf{11.03} & \textbf{41.45}
 & 24.05 & \textbf{39.88} & \textbf{47.35} \\
\hline
\multirow{4}{*}{Mixed datasets} & - & - & - & 3.9 & 11.65 & 48.65 & 20.54 & 38.15 & 53.1 \\
& \checkmark & - & - & 6.1 & 15.65 & 54.18 & 29.55 & 45.23 & 56.59 \\
& \checkmark & \checkmark & - & 7.61 & 16.93 & 53.97 & 36.11 & 46.36 & 55.39\\
 
& \checkmark & \checkmark & \checkmark & \textbf{8.54} & \textbf{17.86} & \textbf{55.08}
 & \textbf{37.29} & \textbf{47.58} & \textbf{57.19} \\
\hline
\end{tabular}}
\end{table}




\begin{figure*}[t!]
    \centering
    \scriptsize
    \setlength{\tabcolsep}{2pt}
    \begin{tabular}{ccccc}
    \centering

        {\includegraphics[width=0.16\linewidth]{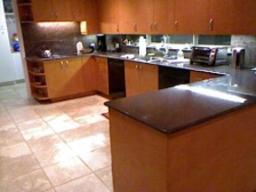}} &  
        {\includegraphics[width=0.16\linewidth]{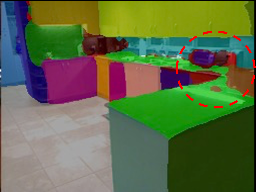}} &  
        {\includegraphics[width=0.16\linewidth]{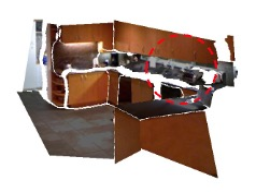}} & 
        {\includegraphics[width=0.16\linewidth]{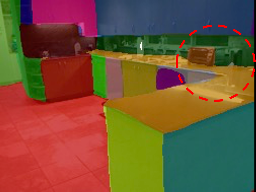}} &
        {\includegraphics[width=0.16\linewidth]{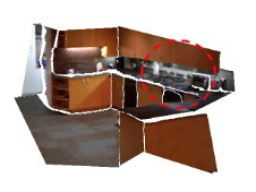}} \\

        {\includegraphics[width=0.16\linewidth]{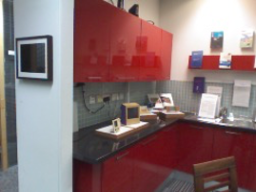}} &  
        {\includegraphics[width=0.16\linewidth]{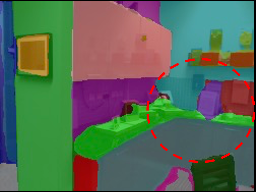}} &  
        {\includegraphics[width=0.16\linewidth]{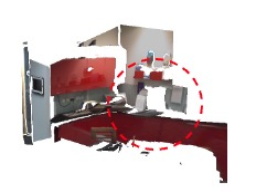}} & 
        {\includegraphics[width=0.16\linewidth]{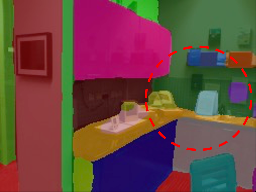}} &
        {\includegraphics[width=0.16\linewidth]{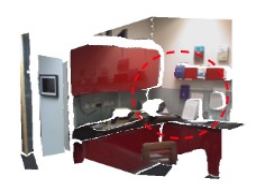}} \\ 


        {\includegraphics[width=0.16\linewidth]{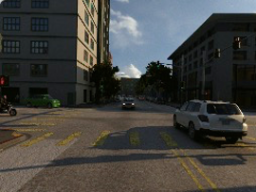}} &  
        {\includegraphics[width=0.16\linewidth]{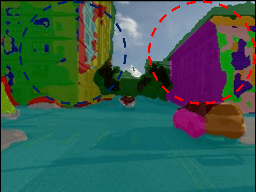}} &  
        {\includegraphics[width=0.16\linewidth]{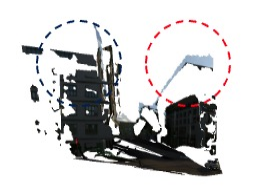}} & 
        {\includegraphics[width=0.16\linewidth]{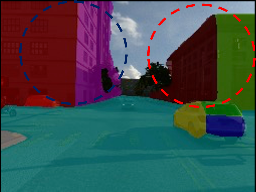}} &
        {\includegraphics[width=0.16\linewidth]{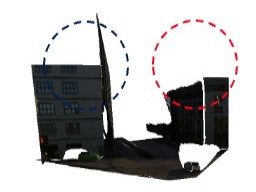}} \\ 

        {\includegraphics[width=0.16\linewidth]{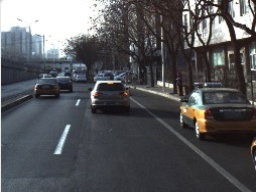}} &  
        {\includegraphics[width=0.16\linewidth]{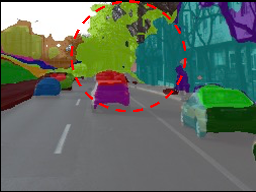}} &  
        {\includegraphics[width=0.16\linewidth]{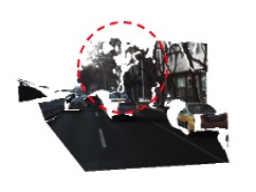}} & 
        {\includegraphics[width=0.16\linewidth]{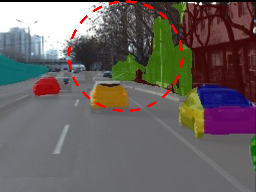}} &
        {\includegraphics[width=0.16\linewidth]{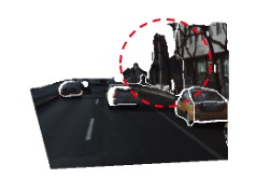}} \\ 
        Image &  PlaneRecTR-Segmentation & PlaneRecTR-Mesh &
        Ours-Segmentation & Ours-Mesh \\
    \end{tabular}
    \caption{Qualitative results from our mix-trained model for zero-shot plane segmentation and mesh reconstruction on NYUv2, 7-Scenes, Parallel Domain, and ApolloScape, from top to bottom, respectively. Noticeable differences are highlighted.}
    \label{fig::qualitative}
\end{figure*}

\noindent\textbf{Training data.} We examine the effect of different training dataset combinations. To verify the effect of training data scale, we construct a `starter' mixed dataset, comprising two indoor datasets~(ScanNetv1, Matterport3D) and two outdoor datasets~(Synthia, VKITTI), then train our model on it. Furthermore, to demonstrate how the domain gap of indoor and outdoor data affects the plane reconstruction, we utilize all of the indoor training datasets and all outdoor datasets respectively, to train models and then conduct zero-shot evaluation on both indoor~(NYUv2) and outdoor~(ParallelDomain) datasets. As suggested in Table~\ref{tab::training_datasets}, the indoor-trained model failed to generalize onto ParallelDomain due to the challenge imposed by the domain gap, and the same holds for using the outdoor-trained model to test on NYUv2. Training with the full mixed dataset yields substantial improvement over the starter dataset, demonstrating the significant benefits of a larger training data scale. Besides, we surprisingly find that our mix-domain model is comparable~(on NYUv2) and even outperforms the single-domain reconstruction result~(on ParallelDomain). This demonstrates the robustness and effectiveness of our system across diverse domains, with strong potential for even better results when incorporating additional training datasets.

\begin{table}[hbp]
\caption{Ablation study on training our framework using different combinations of indoor and outdoor datasets.}
\label{tab::training_datasets}
\centering
\resizebox{1.0\linewidth}{!}{
\begin{tabular}{c|c|ccc|ccc}
\hline
\centering
\multirow{2}{*}{Evaluation Dataset} & \multirow{2}{*}{Training Datasets} & \multicolumn{3}{|c}{Plane Recall (depth)} & \multicolumn{3}{|c}{Plane Recall (normal)} \\
\cline{3-8}
& & @0.05m / 1m & @0.1m / 3m & @0.6m / 10m & @5° & @10° & @30°\\
\hline
\multirow{4}{*}{NYUv2~\cite{silberman2012indoor}} & Indoor Only & \underline{8.41} & \textbf{18.7} & \textbf{55.89} & \textbf{37.82} & \textbf{47.93} & \underline{57.08} \\
& Outdoor Only & 0.43 & 1.71 & 15.59 & 7.9 & 14.68 & 26.75\\
& Starter Mixed~(110K) & 5.24 & 13.43 & 46.26 & 31.2 & 43.14 & 51.04\\ 
& Full Mixed~(560K) & \textbf{8.54} & \underline{17.86} & \underline{55.08} & \underline{37.29} & \underline{47.58} & \textbf{57.19} \\ 
\hline
\multirow{4}{*}{ParallelDomain~\cite{guizilini2021geometric, qian2020learning}} & Outdoor Only & \textbf{26.72} & \underline{43.23} & \underline{59.87} & \underline{48.55} & \underline{59.32} & \underline{64.43}\\
& Indoor Only & 0.81 & 1.19 & 3.66 & 16.94 & 21.4 & 37.57\\
& Starter Mixed~(110K) & 14.04 & 35.45 & 53.53 & 42.54 & 55.91 & 61.36\\ 
& Full Mixed~(560K) & \underline{25.96} & \textbf{44.51} & \textbf{63.45} & \textbf{53.23} & \textbf{65.74} & \textbf{69.62} \\ 
\hline
\end{tabular}
}
\end{table}

\noindent\textbf{Training on high-resolution annotations.} In the former evaluations we report the numbers of training on low-resolution groundtruth~(256x192) for the purpose of fair comparison with the baselines. To illustrate the benefits of training on higher-resolution~(640x480) and our adaptability among different input sizes, in table~\ref{tab::resolution}, we compare the low-res-trained model and high-res-trained model on high-resolution evaluation. One can see that training can higher resolution offers benefits on both plane segmentation and geometry over most metrics. 

\begin{table}[tbp]
\caption{Comparison on training using low vs. high resolution data, then evaluation on high-resolution groundtruth.}
\vspace{-2mm}
\label{tab::resolution}
\centering
\resizebox{1.0\linewidth}{!}{
\begin{tabular}{c|c|ccc|ccc|ccc}
\hline
\centering
\multirow{2}{*}{Evaluation Dataset} & Training  & \multicolumn{3}{c|}{Plane Segmentation} & \multicolumn{3}{|c}{Plane Recall (depth)} & \multicolumn{3}{|c}{Plane Recall (normal)} \\
\cline{3-11}
& Resolution & RI(↑) & VOI(↓)  & SC(↑) & @0.05m / 1m & @0.1m / 3m & @0.6m / 10m & @5° & @10° & @30°\\
\hline
\multirow{2}{*}{NYUv2} & Low-res & \textbf{0.92} & 0.93 & 0.76 & \textbf{8.35} & \textbf{17.78} & 55.33 & 37.41 & 47.87 & 57.35 \\
& High-res & \textbf{0.92} & \textbf{0.84} & \textbf{0.77} & 7.47 & 16.19 & \textbf{57.31} & \textbf{40.09} & \textbf{50.34} & \textbf{59.87} \\
\hline
\multirow{2}{*}{ParallelDomain} & Low-res & \textbf{0.94} & 0.68 & 0.84 & 26.38 & 44.43 & 62.68 & 51.62 & 64.34 & 68.43 \\
& High-res & \textbf{0.94} & \textbf{0.63} & \textbf{0.85} & \textbf{30.38} & \textbf{50.0} & \textbf{66.09} & \textbf{56.55} & \textbf{67.91} & \textbf{71.32} \\
\hline
\end{tabular}
}
\vspace{-10pt}
\end{table}

\noindent\textbf{Qualitative results.} We showcase the plane segmentation and reconstructed meshes on various in-the-wild datasets in Fig.~\ref{fig::qualitative}. Compared to the current SOTA method PlaneRecTR~\cite{shi2023planerectr}, our approach consistently exhibits improvement. As illustrated in various indoor scenes in the top three cases, the reconstructed planes from PlaneRecTR sometimes deviate from the correct geometry, whereas ours more effectively preserve inter-plane geometric relationships such as parallelism or orthogonality. In outdoor scenarios, our method demonstrates significantly improved segmentation and mesh precision, as shown in the last three rows.




%% file: sec/conclusion.tex
\vspace{-5pt}
\section{Conclusion}
\label{sec:conclusion}
In this paper, our primary focus is to establish the pivotal task of mix-domain plane reconstruction from a single image. We generate high-resolution, dense 3D plane labels automatically for several indoor and outdoor datasets, leading to a large-scale, mix-domain benchmark to fit the training requirements. We present a Transformer-based model named ZeroPlane, a unified, transferable 3D plane reconstruction framework, training on mixed domains and datasets. Our model disentangles plane normal and offset representation, and employs an exemplar-guided, classification-then-regression learning scheme and a pixel-geometry-enhanced module to achieve robust and precise plane reconstruction. Extensive experiments have demonstrated the framework's superior zero-shot generalizability on in-the-wild data evaluated across domains. We anticipate that our model and annotated planar benchmark will advance generalization and practical applications, while also inspiring further research in this area. Please find more experimental results and other technical details in our supplementary materials.

\smallskip
\noindent \textbf{Limitations and future work.} Although our system demonstrates strong performance and generalizability, it is still limited by the lack of real outdoor training data due to missing dense ground-truth depth, which partially hinders further improvements in generalization. To make progress, we plan to apply semantic or geometric foundation models such as SAM~\cite{kirillov2023segment} and DepthAnything~\cite{yang2024depth} on unlabeled data and leverage the prior knowledge encoded in these foundation models to address such limitation. 


%% file: sec/X_suppl.tex
\clearpage
\appendix
\setcounter{page}{1}
\maketitlesupplementary

\setcounter{page}{1}

\section{Details on Plane Annotation Generation}
In this section, we present more details about our dense plane annotation generation pipeline on the new benchmark indoor~\cite{straub2019replica, yadav2023habitat, vasiljevic2019diode, zamir2018taskonomy, shotton2013scene} and outdoor datasets~\cite{ros2016synthia, gaidon2016virtual, cabon2020virtual, huang2018apolloscape, waghmare2023sanpo, guizilini2021geometric, qian2020learning}. Figure~\ref{supp_fig::groundtruth} shows examples of our plane annotation on different datasets.

\paragraph{Point cloud lifting.} For RGB-D datasets containing precise ground-truth depth maps, we lift depth map to 3D point cloud for plane fitting. For stereo data such as ApolloScape, we first transform the disparity map into depth map using the provided camera baseline and intrinsic parameters, then lift the depth map and fit planes.
\paragraph{Panoptic segmentation.} For datasets without dense semantic instance ground truth, we employ the state-of-the-art image segmentation approach Mask2Former~\cite{cheng2022masked} to obtain the panoptic segmentation results to assist the plane fitting process. We leverage their released models pretrained on ADE20K~\cite{zhou2017scene} and Cityscapes~\cite{cordts2016cityscapes} to run on our indoor and outdoor datasets, respectively. 
\paragraph{Plane number ranges.} We select the obtained masks from categories likely to contain planar structures into our plane fitting stage, and perform instance-wise plane fitting. Moreover, we empirically set different plane number range~(minimum and maximum number of planes) contained in each mask from either a background stuff or a foreground instance. For instance, for outdoor scenes we set [1, 2] for roads and walls, [1, 5] for buildings, and [0, 2] for vehicles. For indoor scenes, we set [0, 1] for floors and [0, 5] for other furniture. 
\paragraph{Plane fitting with RANSAC.} We follow previous works~\cite{liu2018planenet, liu2019planercnn} to fit planes with RANSAC. Specifically, we run RANSAC for $200$ iterations for each plane. In each iteration, we randomly sample three points from the instance mask to fit a plane hypothesis then compute and record the number of point inliers over the instance point set. We select the plane hypothesis with maximum inliers as the final plane proposal, and use least square algorithm to refit the plane onto the entire set of its inliers and update its parameter. After getting proposals for each instance independently, we merge the neighbouring planes from the same semantic instance if their plane parameters are close to each other. Please refer to the implementation of~\cite{liu2018planenet, liu2019planercnn} for more details.
\paragraph{Distance-aware fitting error thresholds.} Since the geometric scale variation of outdoor data is much larger than that of indoor scenes, we set a more tolerant fitting error~(the average distance of all inlier points to the fitted plane proposal) threshold for the distant points while employing RANSAC. Our motivation is to make the threshold proportional to the average depth of these points. In this way, close and distant points are treated in a roughly equal manner. We set $0.05m$ as the reference fitting error and $10m$ as the reference average depth. Then, the adapted fitting error $E$ of a plane proposal with an average depth $d_m$, is computed as:
\begin{equation}
    E = \max(\frac{0.05*d_m}{10}, 0.05)
\end{equation}
A plane proposal will be rejected from the RANSAC process if its average fitting error exceeds the corresponding error threshold $E$.
\paragraph{Filtering tiny planes.} After RANSAC fitting, we filter out tiny planes (those smaller than $200$ pixels), as they are too challenging to be reliably detected by our annotation model.

\input{supp/figures/supp_groundtruth_figures}

\paragraph{User evaluation on our generated groundtruth.} To intuitively validate the groundtruth quality of our pipeline, we have invited $10$ volunteers to give rating on the plane segmentation quality from $500$ randomly sampled images from all datasets as good, borderline, or bad. We received ratings of 84\% `good', 15\% `borderline', and 1\% `bad', verifying the convincing quality of our generated data.

\paragraph{Limitations on current pipeline.} Although achieving desirable annotation quality over most of the scenes, we acknowledge that our current pipeline still exists a few limitations over some scenarios. First, on real-world data, the depth maps captured by sensors are sometimes incomplete, leading to missing planar mask annotation in our annotation since we leverage the point map lifted by depth. Second, the instance segmentation categories and plane number ranges are pre-defined prior to plane fitting, leading to some undefined regions on some not-well-defined cases. A potential solution is to leverage the SOTA segmentation model such as SAM for open-set segmentation to cover more planes.

\section{Details on Our Method}
\paragraph{Loss Weights.} On the weight coefficients of different loss terms, we empirically set $\lambda_{c}=2.0$ on plane classification, $\lambda_{m}=5.0$ on plane mask for both dice and cross entropy losses, $\lambda_{n_{c}}=1.0$ for normal classification, $\lambda_{n_{r}}=5.0$ for normal residual regression, $\lambda_{d_{c}}=1.0$ for offset classification, $\lambda_{d_{r}}=2.0$ for offset residual regression, $\lambda_{p_{d}}=0.5$, $\lambda_{p_{n\_l1}}=1.0$ for pixel normal $L_{1}$ loss and $\lambda_{p_{n\_cos}}=5.0$ for pixel normal cosine distance loss. 

\paragraph{Network Architecture.} For the use of DINOv2 encoder and DPT pixel decoder, we follow their official implementation. On the pixel depth and normal heads, we feed the pixel features into three consecutive convolutional layers with ReLU activation except for the output layer for depth and normal respectively. For the pixel-geometry enhanced plane embedding module, we first pass the predicted depth and normal separately to a convolutional layer to derive the pixel geometric embeddings, then employ cross-attention, self-attention, and feed-forward network (FFN) between the plane query embeddings and the obtained pixel geometric embeddings to obtain the enhanced plane embeddings. This procedure is similar to the computational manner between query embeddings and pixel features used in query-based transformer detectors, as detailed in  Mask2Former~\cite{cheng2022masked}. Regarding normal and offset classification and residual regression, we use two MLPs which take the instance-level plane embeddings as input and decode the plane class logits and residual vector, respectively. To achieve a better trade-off between precision and computational cost, we decrease every embedding layers dimension used in original~\cite{cheng2022masked} from 256 to 64, where we do not observe a great impact on plane reconstruction performance.

\paragraph{Computational overhead.} We compare our computational overhead with PlaneRecTR~\cite{shi2023planerectr} which shares similar overall architecture with ours. Under our default setting with DINO-B as our encoder, our model has 107.8M parameters and our FLOPS is 285M, whereas PlaneRecTR has 107M parameters and the FLOPS is 265M. We achieve comparable computational cost while significantly better zero-shot generalizability compared with this competitive counterpart.

\section{Additional Experimental Results and Ablation Studies}
In this section, we incorporate more ablation studies to demonstrate the robustness of our model, including the selections of exemplar number, the design of disentangled plane normal and offset used in our system, the robustness of our model on potential data bias, and the employment of SOTA monocular depth estimation with RANSAC as a competitive baseline method.

\paragraph{In-domain evaluation.} Besides zero-shot evaluation, we provide the evaluation results of our model on the validation split of in-domain datasets (ScanNet~\cite{dai2017scannet}, Synthia~\cite{ros2016synthia}) for both single-dataset training and mix-dataset training settings. As shown in Table~\ref{tab::indomain}, in both settings, our method achieves notable improvement for most of the metrics, especially on planar geometry.

\input{supp/tables/supp_table_indomain}

\paragraph{The use of disentangled normal and offset.}

In Tab.\ref{supp_tab::disentangle}, we show the result of an ablation study that compares between without disentanglement (using $n/d$ to represent the plane parameter for classification-then-regression while keeping all the other proposed modules) and with disentanglement. It shows that disentanglement brings remarkable improvements in most of the metrics. This verifies the necessity of applying decoupled representation on normal and offset, whose physically meanings are distinct.

\input{supp/tables/supp_table_disentangle}

\paragraph{The selection of normal and offset exemplar numbers.} We then investigate the impact of varying the number of exemplars on normal and offset in Table~\ref{supp_tab::exemplar}. One can see that, our model is generally robust to the selection of $K_{n}$ and $K_{d}$, where the gaps on different selections are relatively small. Empirically, changing solely normal or offset exemplars does not lead to much gain and our default parameters achieve the best overall performance.
\input{supp/tables/supp_table_exemplar}

\paragraph{Robustness on the source of plane exemplar.} To verify the robustness on how we obtain the clusters of plane normal and offsets on classification-then-regression, we conduct an ablation study by using only 2 indoor and 2 outdoor datasets, as opposed to using all 10 mixed training datasets, for clustering the normal and offset exemplars while still training on the full set of 10 mixed datasets. As shown, although suboptimal clusters led to a marginal performance drop, our model still demonstrated clear robustness over the source of plane examplar clusters.

\input{supp/tables/supp_cluster_source}

\noindent \textbf{Robustness to pixel-level depth and normal prediction.} To validate whether bad pixel depth\&normal prediction can lead to a performance gap on final plane reconstruction, We did an ablation study by adding random Gaussian noise with variation 0.05 w.r.t the original pixel and depth prediction values. As shown in the following table, there are only minor changes, demonstrating the robustness of our framework on depth and normal predictions.
\input{supp/tables/supp_depth_normal_variation}

\paragraph{The bias introduced by Mask2Former~\cite{cheng2022masked} on groundtruth fitting and model design.} One potential concern raised from our proposed plane annotation pipeline and our framework is that, we use Mask2former's panoptic segmentation predictions for instance segmentation then plane fitting during groundtruth generation for a couple of datasets, while our framework is also partially based on Mask2former. This will introduce bias during both training and evaluation especially on the datasets whose groundtruth is involved by Mask2former. To this end, we conduct an ablation experiment, where we use the rest of datasets whose annotation pipeline does not involve Mask2former to train both the baseline counterpart~\cite{shi2023planerectr} and our system, which eliminates the effect brought by Mask2former's involvement on groundtruth labels. As shown in Table~\ref{supp_tab::mask2former_bias}, our method still significantly outperforms the parallel version of PlaneRecTR, which demonstrates the robustness of our model on this potential bias.

\input{supp/tables/supp_table_mask2former_bias}

\input{supp/figures/supp_qualitative}

\paragraph{Employing SOTA monocular depth estimation and segmentation as a competitive baseline.} Inspired by the recent success of foundation models on depth estimation and image segmentation, we apply the SOTA monocular metric depth estimation methods Metric3D-v2~\cite{hu2024metric3d} and DepthPro~\cite{bochkovskii2024depth} to get dense pixel-wise monocular depth, and use Mask2former~\cite{cheng2022masked} for panoptic segmentation. Then, we apply the same RANSAC pipeline as we used on groundtruth plane generation to fit planes. We regard this as a training-free baseline which leverages foundation model inputs to tackle this task. As shown in Table~\ref{supp_tab::monodepth_fitting}, which achieving admissible performance of these two counterparts, we still beat their performance by a large margin, demonstrating our advantage over directly applying foundation models to solve this problem.

\input{supp/tables/supp_monodepth_fitting}

\section{More Qualitative Results}
In Fig.~\ref{supp_fig::qualitative}, we showcase more qualitative results on testing images from diverse benchmarks or newly sampled in-the-wild data. Our model consistently demonstrates effectiveness and robustness across various environments.

%% file: supp/figures/supp_groundtruth_figures.tex
\begin{figure*}[htp]
    \centering
    \scriptsize
    \setlength{\tabcolsep}{2pt}
    \begin{tabular}{ccc}

    \centering
        {\includegraphics[width=0.20\linewidth]{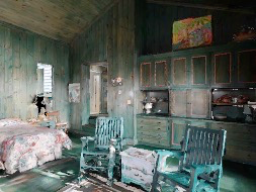}} &  
        {\includegraphics[width=0.20\linewidth]{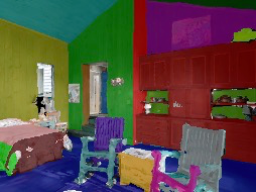}} &  
        {\includegraphics[width=0.20\linewidth]{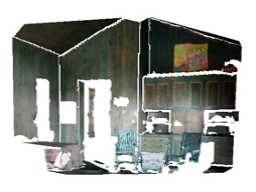}} \\

        {\includegraphics[width=0.20\linewidth]{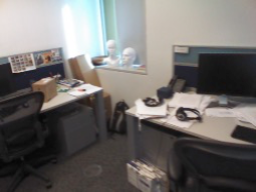}} &  
        {\includegraphics[width=0.20\linewidth]{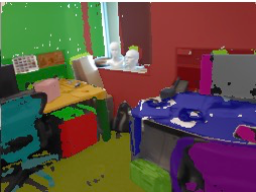}} &  
        {\includegraphics[width=0.20\linewidth]{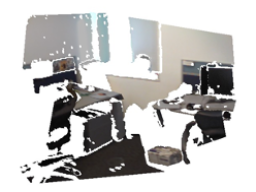}} \\

        {\includegraphics[width=0.20\linewidth]{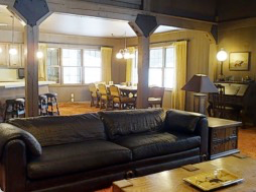}} &  
        {\includegraphics[width=0.20\linewidth]{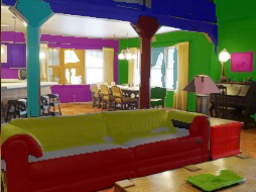}} &  
        {\includegraphics[width=0.20\linewidth]{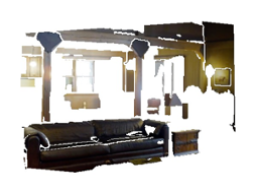}} \\

        {\includegraphics[width=0.20\linewidth]{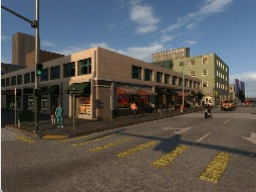}} &  
        {\includegraphics[width=0.20\linewidth]{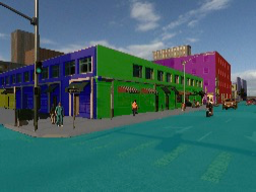}} &  
        {\includegraphics[width=0.20\linewidth]{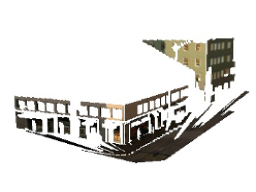}} \\

        {\includegraphics[width=0.20\linewidth]{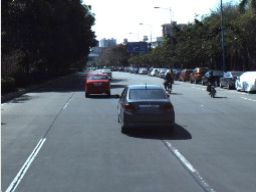}} &  
        {\includegraphics[width=0.20\linewidth]{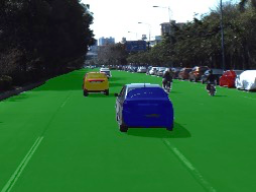}} &  
        {\includegraphics[width=0.20\linewidth]{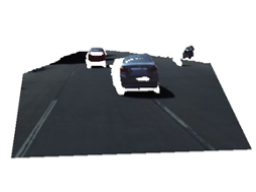}} \\

        {\includegraphics[width=0.20\linewidth]{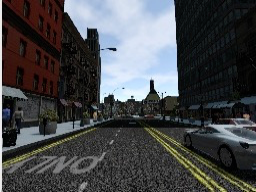}} &  
        {\includegraphics[width=0.20\linewidth]{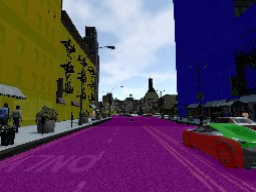}} &  
        {\includegraphics[width=0.20\linewidth]{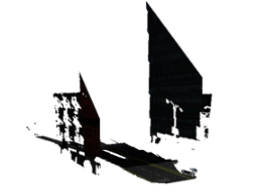}} \\

        {\includegraphics[width=0.20\linewidth]{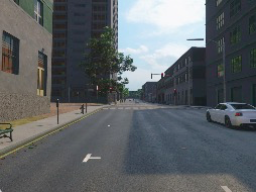}} &  
        {\includegraphics[width=0.20\linewidth]{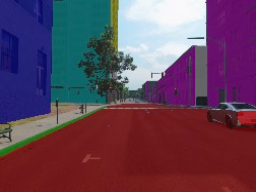}} &  
        {\includegraphics[width=0.20\linewidth]{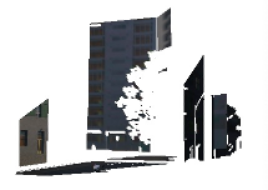}} \\
        
        Image & Segmentation & Mesh \\

    \end{tabular}
    \caption{From top to bottom: our annotated ground-truth planes on HM3D~\cite{yadav2023habitat}, 7-Scenes~\cite{shotton2013scene}, Taskonomy~\cite{zamir2018taskonomy}, ParallelDomain~\cite{guizilini2021geometric, qian2020learning}, ApolloScape~\cite{huang2018apolloscape}, Synthia~\cite{ros2016synthia} and Sanpo~\cite{waghmare2023sanpo} datasets.}
    \label{supp_fig::groundtruth}
\end{figure*}

%% file: supp/tables/supp_table_indomain.tex
\setlength{\tabcolsep}{3pt}
\begin{table}[t!]
\caption{In-domain evaluation of both single-dataset-trained model~(denoted as S) and mix-trained model~(denoted as M) on ScanNet~\cite{dai2017scannet} and Synthia~\cite{ros2016synthia}.}
\label{tab::indomain}
\vspace{-2mm}
\centering
\resizebox{1.00\linewidth}{!}{
\begin{tabular}{c|c|ccc|ccc|ccc}
\hline
\centering
\multirow{2}{*}{Evaluation Dataset} & \multirow{2}{*}{Method} & \multicolumn{3}{c|}{Plane Segmentation} & \multicolumn{3}{c|}{Plane Recall (depth)} & \multicolumn{3}{c}{Plane Recall (normal)} \\ 
\cline{3-11}
 & & RI(↑) & VOI(↓)    & SC(↑)  & @0.05m / 1m     & @0.1m / 3m   & @0.6m / 10m & @5° & @10° & @30°   \\ 
\hline
\multirow{4}{*}{ScanNet~\cite{dai2017scannet}} & \multicolumn{1}{l|}{PlaneRecTR~(S)~\cite{shi2023planerectr}} & \textbf{0.94} & 0.68 & 0.86 & 27.47 & 47.94 & \textbf{77.21} & 49.37 & 65.83 & \textbf{75.24} \\
 & \multicolumn{1}{l|}{Ours~(S)} & \textbf{0.94} & \textbf{0.65} & \textbf{0.87} & \textbf{29.62} & \textbf{48.79} & 74.76 & \textbf{58.18} & \textbf{68.52} & 73.64 \\
\cline{2-11}
 & \multicolumn{1}{l|}{PlaneRecTR~(M)~\cite{shi2023planerectr}} & \textbf{0.91} & \textbf{0.88} & \textbf{0.80} & 18.01 & 37.62 & 75.22 & 37.69 & 59.53 & 72.11\\
 & \multicolumn{1}{l|}{Ours~(M)} & 0.90 & 0.93 & 0.78 & \textbf{21.3} & \textbf{40.43} & \textbf{75.5} & \textbf{55.7} & \textbf{66.78} & \textbf{73.64}\\
\hline
\multirow{4}{*}{Synthia~\cite{ros2016synthia}} & \multicolumn{1}{l|}{PlaneRecTR~(S)~\cite{shi2023planerectr}} & \textbf{0.99} & 0.22 & 0.94 & \textbf{61.52} & 71.32 & 73.80 & 66.46 & 72.87 & 75.37\\
 & \multicolumn{1}{l|}{Ours~(S)} & \textbf{0.99} & \textbf{0.13} & \textbf{0.97} & 61.45 & \textbf{77.04} & \textbf{79.92} & \textbf{79.16} & \textbf{81.37} & \textbf{82.20} \\
\cline{2-11}
 & \multicolumn{1}{l|}{PlaneRecTR~(M)~\cite{shi2023planerectr}} & 0.97 & 0.50
& 0.87 & 40.85 & 50.44 & 57.38 & 41.62 & 52.84 & 59.54\\
 & \multicolumn{1}{l|}{Ours~(M)} & \textbf{0.99} & \textbf{0.17} & 
\textbf{0.96} & \textbf{49.49} & \textbf{62.61} & \textbf{71.23} & \textbf{67.89} & \textbf{72.48} & \textbf{73.66} \\
\hline
\end{tabular}}
\end{table}

%% file: supp/tables/supp_table_disentangle.tex
\begin{table}[t!]
\centering
\caption{Quantitative results on employing coupled or disentangled plane normal and offset on NYUv2~\cite{silberman2012indoor} dataset.}
\resizebox{0.95\linewidth}{!}{
\begin{tabular}{cccccccc}
\toprule
\centering
\multirow{2}{*}{Settings} & \multicolumn{3}{c}{Plane Recall (depth)} & \multicolumn{3}{c}{Plane Recall (normal)} \\ 
\cmidrule(r){2-7} 
 & @0.05m     & @0.1m   & @0.6m & @5° & @10° & @30°   \\ 
\midrule 
Coupled normal and offset & 7.9 & \textbf{17.94} & \textbf{55.76} & 34.48 & 46.63 & 56.61 \\
Disentangled normal and offset & \textbf{8.54} & 17.86 & 55.08 & \textbf{37.29} & \textbf{47.58} & \textbf{57.19} \\
\bottomrule
\end{tabular}}
\label{supp_tab::disentangle}
\end{table}

%% file: supp/tables/supp_table_exemplar.tex
\begin{table}[t!]
\centering
\caption{Quantitative results on employing different numbers of normal and offset exemplars on NYUv2~\cite{silberman2012indoor} dataset.}
\resizebox{0.95\linewidth}{!}{
\begin{tabular}{cccccccc}
\toprule
\centering
\multirow{2}{*}{Settings} & \multicolumn{3}{c}{Plane Recall (depth)} & \multicolumn{3}{c}{Plane Recall (normal)} \\ 
\cmidrule(r){2-7} 
 & @0.05m     & @0.1m   & @0.6m & @5° & @10° & @30°   \\ 
\midrule 
$K_{n}=14, K_{d}=20$ & 8.27 & \textbf{17.98} & 54.9 & 36.46 & 47.18 & 56.67 \\
$K_{n}=7, K_{d}=10$ & 8.21 & 17.84 & 54.67 & 36.71 & 47.47 & 56.67 \\
$K_{n}=7, K_{d}=20$~(our default setting) & \textbf{8.54} & 17.86 & \textbf{55.08} & \textbf{37.29} & \textbf{47.58} & \textbf{57.19} \\
\bottomrule
\end{tabular}}
\label{supp_tab::exemplar}
\end{table}

%% file: supp/tables/supp_cluster_source.tex
\begin{table}[h]
\centering
\resizebox{1.0\linewidth}{!}{
\begin{tabular}{c|c|ccc|ccc}
\hline
\centering
Evaluation Dataset & Cluster source & @0.05m & @0.1m & @0.6m & @5° & @10° & @30° \\
\hline
\multirow{2}{*}{NYUv2} & partial~(4 datasets) & 7.73 & 17.2 & 54.59 & 37.02 & 47.56 & 56.28 \\
& full~(10 datasets) & \textbf{8.54} & \textbf{17.86} & \textbf{55.08} & \textbf{37.29} & \textbf{47.58} & \textbf{57.19} \\

\hline
\end{tabular}
}
\end{table}

%% file: supp/tables/supp_depth_normal_variation.tex
\begin{table}[h]
\centering
\resizebox{1.0\linewidth}{!}{
\begin{tabular}{c|c|ccc|ccc}
\hline
\centering
Evaluation Dataset & Pixel depth \& normal & @0.05m & @0.1m  & @0.6m & @5° & @10° & @30° \\
\hline
\multirow{2}{*}{NYUv2} & Adding noise & 8.52 & 17.88 & 55.04 & 37.31 & 47.62 & 57.21 \\
& Model Prediction & 8.54 & 17.86 & 55.08 & 37.29 & 47.58 & 57.19 \\
\hline
\end{tabular}
}
\end{table}

%% file: supp/tables/supp_table_mask2former_bias.tex
\begin{table}[t!]
\centering
\caption{Quantitative results on training without Mask2former-produced datasets, then evaluating on NYUv2.}
\resizebox{0.95\linewidth}{!}{
\begin{tabular}{cccccccc}
\toprule
\centering
\multirow{2}{*}{Settings} & \multicolumn{3}{c}{Plane Recall (depth)} & \multicolumn{3}{c}{Plane Recall (normal)} \\ 
\cmidrule(r){2-7} 
 & @0.05m     & @0.1m   & @0.6m & @5° & @10° & @30°   \\ 
\midrule 
PlaneRecTR w/o Mask2former data & 5.01 & 13.47 & 49.29 & 19.16 & 36.69 & 50.77 \\
Ours w/o Mask2former data & \textbf{7.22} & \textbf{16.37} & \textbf{49.8} & \textbf{33.66} & \textbf{43.68} & \textbf{52.05} \\
\bottomrule
\end{tabular}}
\label{supp_tab::mask2former_bias}
\end{table}

%% file: supp/figures/supp_qualitative.tex
\begin{figure*}[htp]
    \centering
    \scriptsize
    \setlength{\tabcolsep}{2pt}
    \begin{tabular}{ccc}

    \centering
        {\includegraphics[width=0.20\linewidth]{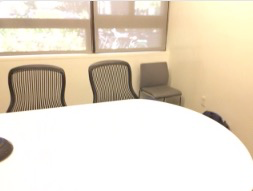}} &  
        {\includegraphics[width=0.20\linewidth]{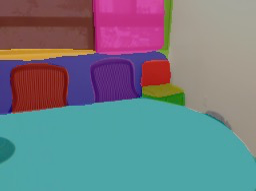}} &  
        {\includegraphics[width=0.20\linewidth]{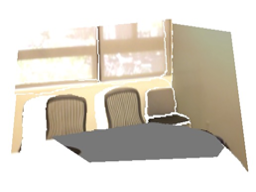}} \\

        {\includegraphics[width=0.20\linewidth]{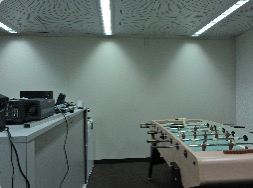}} &  
        {\includegraphics[width=0.20\linewidth]{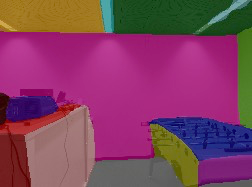}} &  
        {\includegraphics[width=0.20\linewidth]{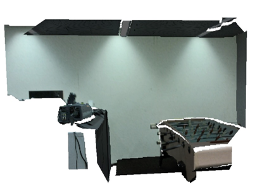}} \\

        {\includegraphics[width=0.20\linewidth]{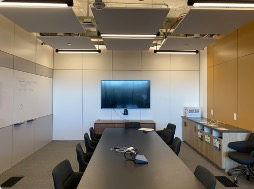}} &  
        {\includegraphics[width=0.20\linewidth]{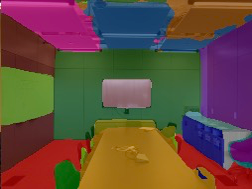}} &  
        {\includegraphics[width=0.20\linewidth]{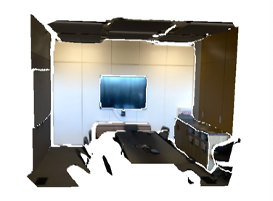}} \\

        {\includegraphics[width=0.20\linewidth]{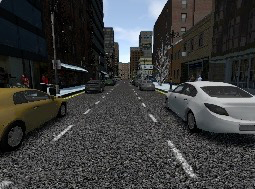}} &  
        {\includegraphics[width=0.20\linewidth]{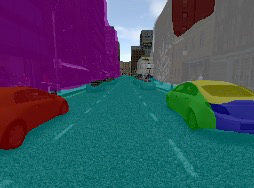}} &  
        {\includegraphics[width=0.20\linewidth]{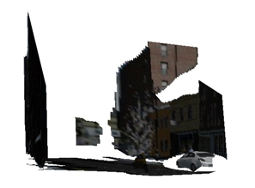}} \\

        {\includegraphics[width=0.20\linewidth]{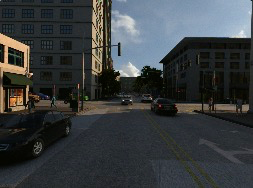}} &  
        {\includegraphics[width=0.20\linewidth]{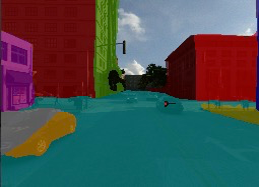}} &  
        {\includegraphics[width=0.20\linewidth]{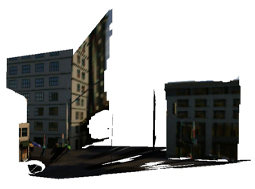}} \\

        {\includegraphics[width=0.20\linewidth]{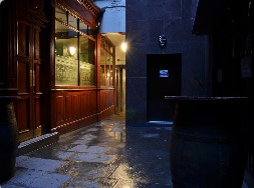}} &  
        {\includegraphics[width=0.20\linewidth]{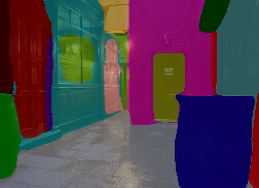}} &  
        {\includegraphics[width=0.20\linewidth]{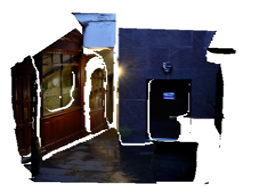}} \\

        {\includegraphics[width=0.20\linewidth]{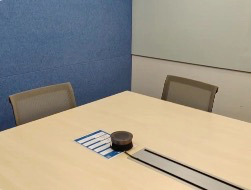}} &  
        {\includegraphics[width=0.20\linewidth]{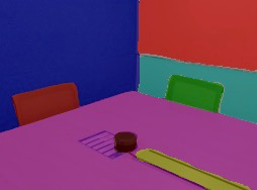}} &  
        {\includegraphics[width=0.20\linewidth]{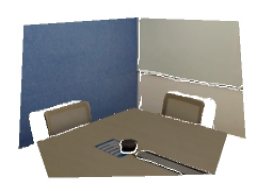}} \\

        {\includegraphics[width=0.20\linewidth]{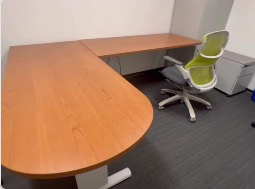}} &  
        {\includegraphics[width=0.20\linewidth]{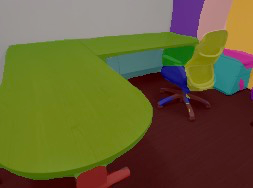}} &  
        {\includegraphics[width=0.20\linewidth]{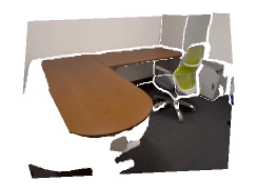}} \\

        Image & Segmentation & Mesh \\

    \end{tabular}
    \caption{From top to bottom: the plane segmentation and reconstruction visualization of our model on ScanNet~\cite{dai2017scannet}, ETH3D~\cite{schops2017multi}, LLFF~\cite{mildenhall2019local}, Synthia~\cite{ros2016synthia}, ParallelDomain~\cite{guizilini2021geometric, qian2020learning}, OASIS~\cite{chen2020oasis} and two in-the-wild images captured by ourselves.}
    \label{supp_fig::qualitative}
\end{figure*}

%% file: supp/tables/supp_monodepth_fitting.tex
\begin{table}[t!]
\centering
\caption{Quantitative results on training without Mask2former-produced datasets, then evaluating on NYUv2.}
\resizebox{0.95\linewidth}{!}{
\begin{tabular}{cccccccc}
\toprule
\centering
\multirow{2}{*}{Settings} & \multicolumn{3}{c}{Plane Recall (depth)} & \multicolumn{3}{c}{Plane Recall (normal)} \\ 
\cmidrule(r){2-7} 
 & @0.05m     & @0.1m   & @0.6m & @5° & @10° & @30°   \\ 
\midrule 
Metric3D + Mask2former + RANSAC & 2.72 & 6.76 & 47.02 & 14.09 & 34.11 & 47.56 \\
Depth-Pro + Mask2former + RANSAC & 3.61 & 9.14 & 47.91 & 20.11 & 37.41 & 49.52 \\
Ours & \textbf{8.54} & \textbf{17.86} & \textbf{55.08} & \textbf{37.29} & \textbf{47.58} & \textbf{57.19} \\
\bottomrule
\end{tabular}}
\label{supp_tab::monodepth_fitting}
\end{table}